\documentclass[10pt,a4paper,twocolumn]{NARMS}

\usepackage{timesmt}
\usepackage{caption}
\usepackage[backend=biber, sorting=none, style=ieee]{biblatex}
\usepackage{hyperref}
\usepackage{bbm}

\usepackage{url}
\usepackage{array}
\usepackage{pgfplots} 
\pgfplotsset{compat=newest} 
\usepackage{multirow}
\usepackage{amssymb}
\usepackage{amsmath}
\usepackage{graphicx}
\usepackage{color}
\usepackage[english]{babel}
\usepackage{verbatim}
\usepackage{tikz}
\usepackage{booktabs}
\usepackage{mathtools}
\usepackage{siunitx}
\usepackage{accents}

\usepackage[utf8]{inputenc}
\usepackage[english]{babel}
\usepackage{booktabs}
\usepackage{balance}

\usepackage{amsthm}

\usepackage[utf8]{inputenc}
\usepackage{algorithm}
\usepackage{esvect}
\usepackage[noend]{algpseudocode}

\graphicspath{{figures/}}
\renewcommand{\phi}{\varphi}
\newcommand{\norm}[1]{\left\lVert#1\right\rVert}

\confshortname{AVEC'18}
	
\title{Hardware-In-the-Loop for Connected Automated Vehicles Testing in Real Traffic}
\author{{Yeojun Kim$^{\ast}$, Samuel Tay$^{\ast}$, Jacopo Guanetti, Francesco Borrelli} \\
{\aff{Department of Mechanical Engineering}} \\
{\aff{University of California, Berkeley, USA}} \\
\\
{\authornext{Ryan Miller}}\\
{\aff{Hyundai Motor Group}} \\
{\aff{\normalfont\normalsize E-mail: yk4938@berkeley.edu}}\\
{\aff{\normalfont\normalsize Topics/hardware-in-the-loop, simulation, connected automated vehicles, model predictive control}}
}

\abstract{
We present a hardware-in-the-loop (HIL) simulation setup for repeatable testing of Connected Automated Vehicles (CAVs) in dynamic, real-world scenarios.
Our goal is to test control and planning algorithms and their distributed implementation on the vehicle hardware and, possibly, in the cloud.
The HIL setup combines PreScan for perception sensors, road topography, and signalized intersections; Vissim for traffic micro-simulation; ETAS DESK-LABCAR/a dynamometer for vehicle and powertrain dynamics; and on-board electronic control units for CAV real time control.
Models of traffic and signalized intersections are driven by real-world measurements.
To demonstrate this HIL simulation setup, we test a Model Predictive Control approach for maximizing energy efficiency of CAVs in urban environments.
}

\begin{document}
\maketitle

\section{INTRODUCTION}\label{sec:Introduction}
\let\thefootnote\relax\footnote{$^{\ast}$These authors equally contributed to this work.}
Recent advances in sensing and connectivity technologies have enabled the development of Connected Automated Vehicles (CAVs).
In particular, CAVs have been studied extensively for their potential to improve vehicle safety, traffic flow, and energy efficiency \cite{guanetti2018control}.
In order to objectively validate safety and performance improvements, CAVs require extensive testing under complex traffic scenarios in order to reproduce as closely as possible real world driving conditions. 
These testing scenarios can be challenging, expensive, unsafe, and are often impossible to reproduce consistently.

Many of these issues and risks are often encountered in the development of automotive control systems and control systems in general; to mitigate these issues and risks  Hardware-in-the-loop (HIL) simulations are used, where the processors and parts of the plant are physical, while the rest of the system is simulated by real-time software \cite{bacic2005hardware, fathy2006review}. 
HIL setups are routinely used in industrial practice; in the automotive field, HIL simulations are a common step in the development of Electronic Control Units (ECUs) for various applications, including engine control, suspension control, powertrain control \cite{isermann1999hardware, choi2000sliding,baracos2001enabling}. 

Recently, some HIL setups have been developed to test CAVs in realistic traffic conditions.
In \cite{montanaro2014novel}, HIL simulations were used to validate the stability of a decentralized cooperative adaptive cruise control for platoons.
In \cite{engelbrecht1999development, wells2001hardware}, HIL setups were developed for realistic traffic simulation, but these simulations were only used for testing traffic controllers.
In \cite{tideman2013simulation}, iterative simulations between the vehicle dynamics and the traffic simulator were required to evaluate the vehicle controller performance and the effect of the vehicle on traffic flow.
However, these setups do not consider the simultaneous simulation of CAVs and data-based traffic flow, where both the CAVs and other traffic interact among themselves.

Other studies propose a concurrent simulation of a vehicle's dynamics and its surrounding environment. 
In \cite{short2005hardware}, surrounding vehicles are simulated using a a simplified
dynamics model; in \cite{ma2013hybrid}, the Academy of Military Transportation Simulator (AMTS), which consists of scaled vehicles running on a simulated road, was built for testing cooperative driving of a vehicle platoon; in \cite{gietelink2006development,gietelink2004pre}, experiments with full-scale intelligent vehicles were conducted on a Vehicle Hardware-in-the-Loop (VEHIL) simulator, which is an indoor laboratory facility where the relative motions between the test vehicle and obstacles are reproduced while the test vehicle is placed on a chassis dynamometer (dyno).
These HIL setups focus on the interactions between a few nearby vehicles, rather than on the interaction of one or more CAVs with real-world traffic, simulated based on data, on a complex road network.

Our paper introduces a hardware-in-the-loop setup which can overcome these issues.
In particular, we present an HIL platform for CAV control and planning algorithms testing and validation in complex environments; we use an ETAS DESK-LABCAR developed with real vehicle measurements for high fidelity vehicle dynamics and powertrain simulation.
Our control, planning, estimation, and environment prediction algorithms are distributed between the on-board ECUs (such as a dSpace MicroAutoBox and Matrix embedded PC-Adlink) and the cloud (such as Amazon Web Services).
The rest of the system is simulated in real-time on a desktop machine using PreScan and PTV Vissim.
PreScan is a simulation platform that contains high fidelity models of vehicle dynamics, perception sensors, and the environment (including road topology and signalized intersections) \cite{Prescan}.
PTV Vissim is a microscopic multi-modal traffic flow simulation software \cite{Vissim}.
We leverage traffic and intersection phase data to generate intersections and micro-traffic simulations in PreScan and Vissim.
The software above is integrated with ETAS DESKTOP-LABSCAR and the on-board ECUs enabling interactive simulations, where CAVs react to the surrounding traffic and \emph{vice versa}.

The main benefits of our HIL setup are threefold. First, we exploit advanced simulations in CAVs and complex traffic simulations which are modeled with experimental and collected data. Second, our externally controlled virtual vehicle and other vehicles in the traffic network are mutually interactive in the simulation environment. Lastly, we introduce the HIL design capable of vehicle-to-everything (V2X) communication and cloud connectivity.

We demonstrate our HIL by applying the recently developed eco-friendly adaptive cruise control, designed for energy consumption reduction (eco-ACC) \cite{turri2017model}. 
In eco-ACC, a Model Predictive Control framework is used to avoid a collision while seeking to minimize braking using velocity prediction of the front vehicle.
Our HIL simulations with eco-ACC are conducted in two different scenarios: in the first scenario, a test vehicle drives down a virtual road with a front vehicle running at a constant velocity and sending its velocity trajectory to the following test vehicle. 
In the second scenario, our test vehicle drives through a virtual urban road which consists of other traffic vehicles and traffic lights.
With these demonstrations, we show the capability of our HIL setup for CAVs testing and validation.

The remainder of this paper is organized as follows. Section 2 introduces our HIL hardware and software architecture and explains each component in our setup. Section 3 shows our HIL demonstration with eco-ACC. Section 4 concludes the work with some remarks and possible future work.

\section{HIL Setup} \label{sec:hilformulation}
Our goal for building an HIL simulator is to accurately and fairly validate the on-board ECUs for improving energy efficiency of CAVs.
In order to achieve this, our HIL setup must satisfy the following requirements. First, the environment must be easily constructed and consistently reproducible. Second, the vehicle and powertrain dynamics and their measurements must be consistent with those of a real vehicle. Third,  environment and test vehicle must interact with each other.

In this section we discuss the hardware and software components of our HIL architecture to meet the requirements addressed above. 

\subsection{HIL Hardware Setup} \label{sec:hil_hardware}

\begin{figure}[!h]
	\centering
 	\includegraphics[width=\linewidth,height=\textheight,keepaspectratio]{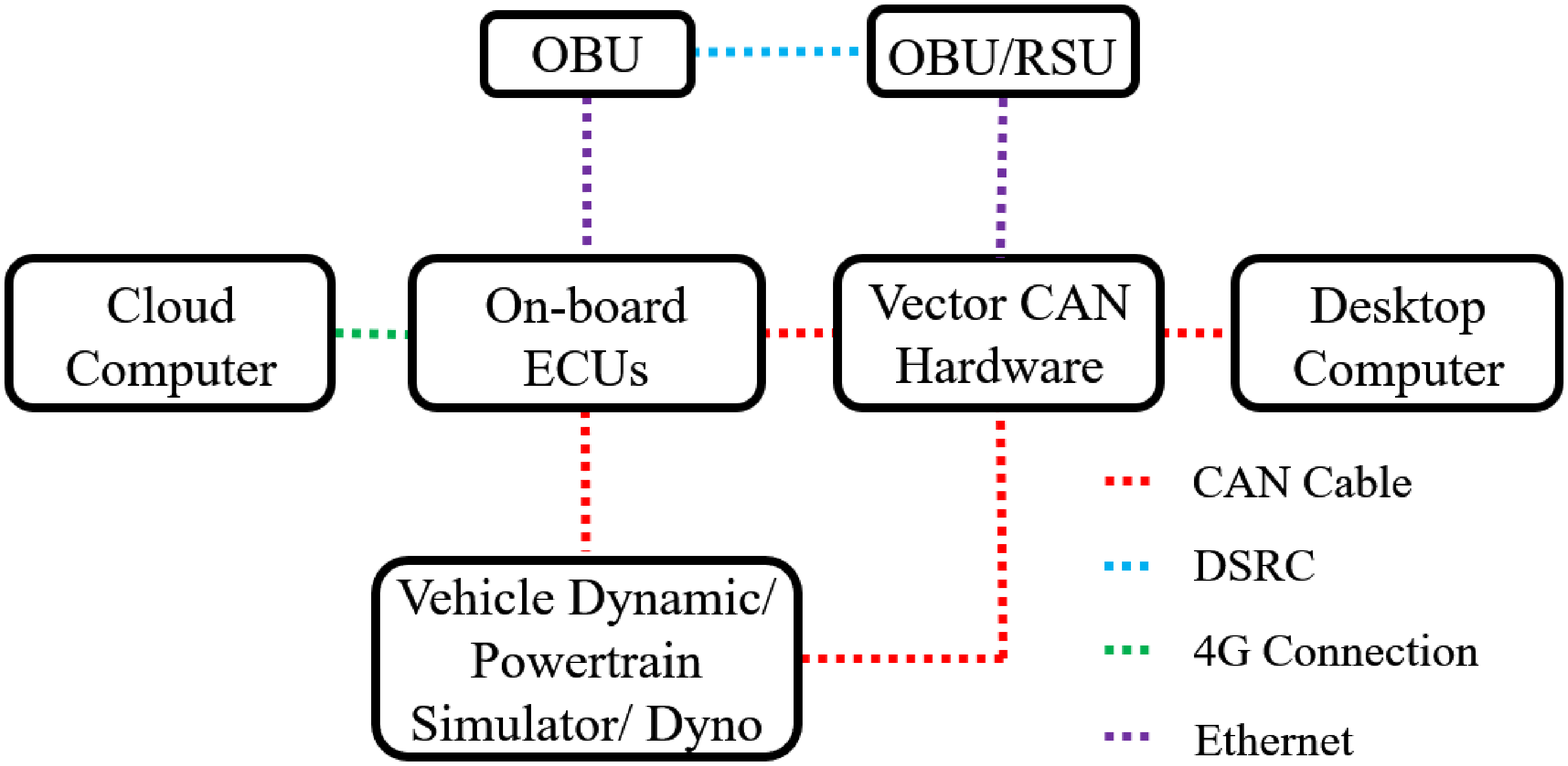}
	\caption{HIL hardware setup schematic}
	\label{fig:hardware_schematic}
\end{figure}

HIL can include different hardware components depending on the availability of their high-fidelity models. 
In our HIL setup, we replace a test vehicle and environment with simulators while the on-board ECUs with connectivity are the same as in the real experiment.
As depicted in Figure~\ref{fig:hardware_schematic}, our HIL hardware setup mainly consists of a desktop computer, a dedicated vehicle dynamic and powertrain simulator/Dyno, and on-board ECUs.
Communications among the desktop computer, the on-board ECUs, and a vehicle dynamic/powertrain simulator happen over CAN bus as in the actual vehicle.
We use a Vector CAN interface to link a desktop computer to the CAN bus.

A desktop computer runs the real-time environment simulation software and mirrors a test vehicle from the real-time vehicle data from a dedicated vehicle dynamic/powertrain simulator. Our desktop has an Intel(R) Core(TM) i7-7700KK CPU @ 4.20Hz with NVIDIA GeForce GTX 1080.

CAV control and planning algorithms exploit a large amount of data, such as sensor data, historical data and look-ahead information about the downstream road; many algorithms are based on optimal control principles, such as Model Predictive Control, which can demand high computational power. To mitigate this issue, the on-board ECUs, including a dSpace MicroAutoBox (IBM PowerPC 750FX processor, 800 MHz) and a Matrix embedded PC-Adlink (MXC-6101D/M4G with Intel Core i7-620LE 2.0 GHz processor), along with its connectivity with cloud computing service, can distribute the computational load appropriately. 
For more details about their specifications and tasks we refer to \cite{guanetti2018control}. In this manner, the optimal controller and the necessary data processing are implemented and executed in real time on our ECUs.

Our ECUs also communicate with other CAVs and infrastructure using an On-board Unit (OBU) and Roadside Unit (RSU), respectively. For these units, we use a Cohda MK5, which supports Dedicated Short Range Communication (DSRC). Communicated messages are defined by the on-board ECUs (for the test vehicle) and the desktop computer (for other participating vehicles and infrastructure).

For a dedicated vehicle dynamic and powertrain simulator, we use an ETAS LABCAR RTPC (ES5100.1 with Intel Core i7-4770S@3.1GHz and 4 CAN Bus interface), allowing for the high-fidelity vehicle models in our HIL simulation. Furthermore, because the interface between the ETAS LABCAR RTPC and the on-board ECUs is established through the CAN bus, we can simply replace the ETAS LABCAR RTPC with a real vehicle on a dynamometer(Dyno) for more accurate measurements of a vehicle and its dynamics.

\begin{figure}[!h]
	\centering
 	\includegraphics[width=\linewidth,height=\textheight,keepaspectratio]{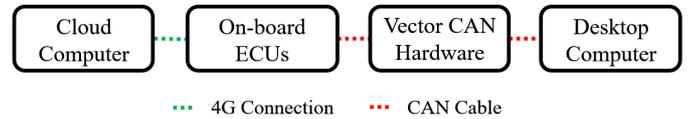}
	\caption{Simplified HIL hardware setup schematic}
	\label{fig:hardware_schematic_simplified}
\end{figure}

It is also noted that this HIL setup can be further simplified by replacing the ETAS LABCAR RTPC, OBU, and RSU with the desktop simulator using Simulink models of vehicle dynamics/powertrain and DSRC modules offered by PreScan. Figure~\ref{fig:hardware_schematic_simplified} depicts the simplified HIL hardware setup.

\subsection{HIL Software Setup} \label{sec:hil_software}
\begin{figure}[!h]
	\centering
 	\includegraphics[width=\linewidth,height=\textheight,keepaspectratio]{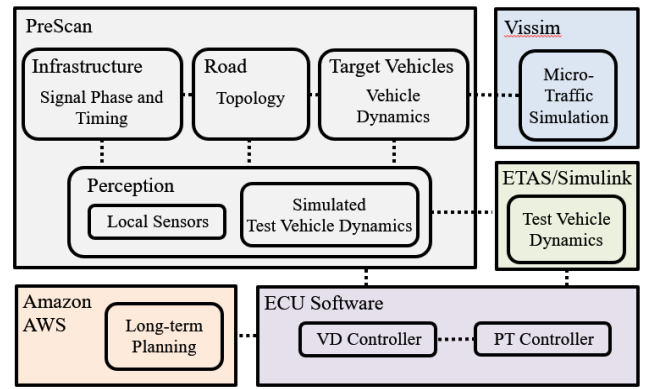}
	\caption{HIL software setup schematic}
	\label{fig:software_schematic}
\end{figure}

High fidelity environment and/or vehicle simulators are employed to represent various complex real world scenarios and test vehicle dynamics. 
Figure \ref{fig:software_schematic} depicts a schematic of the HIL software, which mainly consists of the PreScan and Vissim simulators, ECU software, Amazon AWS, and a vehicle dynamics simulator.
It should be noted that only PreScan and Vissim are running in the desktop computer while other parts are operating on their own designated hardware from Figure~\ref{fig:hardware_schematic}.

\subsubsection{Environment} \label{sec:envir_sim}

The environment simulation models the roads, traffic infrastructure, vehicles in the traffic network, and other external influences from the environment. The PreScan software serves as the main software to build the environment and also as an interface among all other software. For more details on how to build virtual environments using PreScan, we refer to \cite{Prescan}. 
PreScan also includes a built-in plugin for Vissim integration that enables microscopic traffic simulation within the PreScan simulation environment \cite{Vissim}.
PreScan and Vissim are synchronized and have identical road networks.
In order to model traffic flow realistically, we utilize collected traffic data, which provide aggregate traffic volume and turn counts at each intersection, as well as raw traffic signal data.
Within Vissim, vehicles are injected into the road network and make turns at each intersection according to a probabilistic turn policy based on this data. 

\subsubsection{Vehicle and Powertrain Dynamics} \label{sec:vehicle_sim}
We use either Dyno or ETAS software (with the hardware setup in Figure~\ref{fig:hardware_schematic}) or Simulink (with the hardware setup in Figure~\ref{fig:hardware_schematic_simplified}) to model the test vehicle and powertrain dynamics, receiving the control input from the ECUs.
Then, Prescan mirrors the test vehicle dynamics based on the measurements from ETAS software (or Simulink) in the virtual environment.
Therefore, in the PreScan virtual environment, the other traffic target vehicles react to the mirrored test vehicle.
In return, the perception sensors, equipped on the mirrored test vehicle in the PreScan virtual environment, offer the real-time sensor data to our ECUs.

\subsubsection{Controller} \label{sec:cont_sim}
Our controller is implemented in the ECUs, which use dSPACE ControlDesk (in a dSpace MicroAutoBox) and Robotic Operating System (in a Matrix embedded PC-Adlink).
Our ECUs receive, in real time, the vehicle and perception sensor CAN data from ETAS software and PreScan, respectively. 
This data set is the same as the type of data which the on-board ECUs receive during real experiments.
Moreover, our ECUs can be linked to the cloud computing which uses Amazon AWS and calculates a real-time long-term plan for the vehicle.  
Using these real-time data, our ECUs compute the vehicle dynamic/powertrain control action and send it to the vehicle simulator in ETAS software.

In order to make our HIL simulation reflect the real-world in terms of operating frequency and system delay, we first ensure that the PreScan and Vissim simulator in the desktop computer can run at a real-time frequency. Also the ETAS-desktop communication should run at a frequency greater than the frequency of the mirrored vehicle dynamics emulated(mirrored) in the desktop PreScan simulator.

\section{Model Predictive Control in HIL}\label{sec:HIL_testing}

In this section, we test in the proposed HIL simulator the eco-ACC approach on the test vehicle (hereafter referred to as the \emph{ego vehicle}) using Model Predictive Control (MPC) from \cite{turri2017model}.
The MPC controller exploits the advantages brought by CAV technologies to minimize energy consumption. 
This includes the velocity forecast of a front vehicle, Signal Phase and Timing (SPaT), and downstream road traffic flow (enabled by V2X communication). 
Because MPC uses models of the environment to make predictions, it is critical to have realistic dynamics of both the environment and the vehicle itself. 
Therefore, our HIL is a suitable design to test the eco-ACC using MPC.

\subsection{Car following eco-ACC}
In a car following eco-ACC we compute the optimal input trajectories $u^{\ast}(\cdot|t)$ by solving at time $t$ the following finite horizon optimization problem:
\begin{subequations}
\begin{align}
\underset{u(\cdot|t)}{\text{argmin}} \quad &J = 
\sum_{k=t}^{t+N_p} \norm{v (k | t) - v_\text{des}}_{Q} \label{eq:distpenalty}\\
 &\;\;\,+\sum_{k=t}^{t+N_p-1} \norm{F_{b} (k | t)}_{B} \label{eq:brakepenalty}\\
\text{subject to} \quad 
&x(k+1|t) = f(x(k|t),\, u(k|t),\, v_p(k))\,,\label{eq:predmodel}\\
&d_\text{min} \leq d (k|t) \label{eq:safety}\,,\\
&v_\text{min} \leq v (k|t) \leq v_\text{max}\,, \\
&0 \leq F_{t} (k|t) \leq F_{\text{max}}\,, \\
&F_{\text{min}} \leq F_{b} (k|t) \leq 0\,, \\
&x(t|t) = x(t)\,, \label{eq:initialcondition}\\
&\forall k=t,...,t+N_p-1, \nonumber \\
&\begin{bmatrix}d(t+N_p|t)\\v(t+N_p|t)\end{bmatrix} \in \mathcal{C}\,,
\label{eq:terminalset}
\end{align}
\label{eq:mpcproblem}
\end{subequations}
where $x(k|t)=[d(k|t),\, v(k|t),\, F(k|t)]^\text{T}$ and $u(k|t)=[F_t(k|t),\, F_b(k|t)]^\text{T}$ represent the state (which includes distance to a front car $d$, velocity $v$, wheel force $F$) and input (which includes desired traction force $F_t$ and braking force $F_b$) at time $k$ predicted at time $t$, respectively. $N_p$ denotes the prediction horizon of our problem.

The cost function $J$ includes a penalty for tracking the desired velocity $v_\text{des}$ \eqref{eq:distpenalty} and a penalty on braking \eqref{eq:brakepenalty}. 
\eqref{eq:predmodel} represents the vehicle prediction model defined in \cite{turri2017model} where $v_p(k)$ is the predicted velocity of the front vehicle at time $k$.
(1d) represents the safety distance constraint to avoid collisions with the front vehicle; 
(1e) represents the velocity bounds;
(1f) and (1g) represent the input constraint, enforcing the physical limitations of the vehicle and its actuators.
\eqref{eq:initialcondition} represents the initial condition.
Finally, \eqref{eq:terminalset} enforces the terminal state to lie inside the \textit{coasting} set defined in \cite{turri2017model}.

We apply the first input $u^{*,i}_{0|t}$ is applied to the system during the time interval $[t,t+1)$ and at the next time step $t+1$, a new optimal control problem \eqref{eq:mpcproblem} with new measurements of the state, is solved over a shifted horizon, yielding a \textit{moving} or \textit{receding} horizon control strategy with control law.

For the HIL simulation we reproduce the catch-up of a vehicle running at a constant velocity.
The HIL hardware setting depicted in Figure~\ref{fig:hardware_schematic} is employed with the high fidelity model implemented in ETAS LABCAR RTPC. 
Also, in the virtual environment, the ego vehicle is equipped with a Radar sensor modeled by PreScan and receives the perfect velocity prediction of a front vehicle through the Simulink interface. 
The main objective of this simulation is that we can test a safe and fuel efficient eco-ACC with \eqref{eq:mpcproblem} in our HIL environment.

\begin{figure}[!h]
	\centering
 	\includegraphics[width=\linewidth,height=\textheight,keepaspectratio]{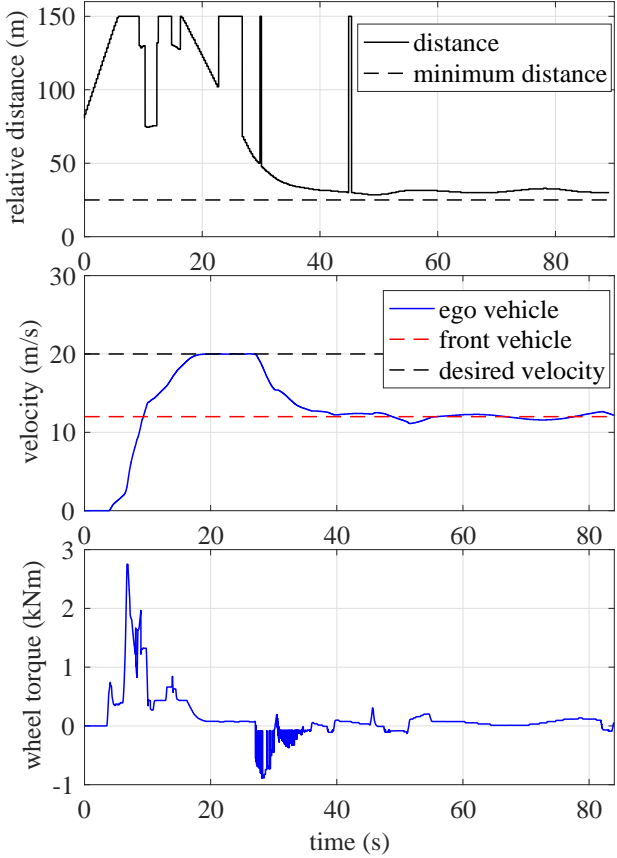}
	\caption{HIL simulation for catch-up of a vehicle traveling with constant velocity profile: inter-vehicle distance, velocity, wheel torque}
	\label{fig:ecoacc}
\end{figure}

Figure~\ref{fig:ecoacc} depicts the closed loop trajectories corresponding to relative distance, velocity and estimated wheel torque of the ego vehicle.
When the target vehicle (the front vehicle) is lost from the PreScan Radar sensor, the distance jumps to $150$m.
As shown, the ego vehicle at first accelerates to reach its desired velocity. Then, when the distance to the front vehicle reaches close to the minimum safety distance, our controller applies small braking torques and maintains its velocity close to the front vehicle velocity. 
Although our eco-ACC is designed to minimize braking, our controller still applies a small braking due to the model mismatch between the high fidelity model in ETAS software and the control-oriented model used in the MPC formulation for eco-ACC.

\subsection{Eco-ACC in Urban Environment}
For eco-ACC in urban environment we modify the optimization problem \eqref{eq:mpcproblem} by adding to our system dynamics in \eqref{eq:predmodel} a state $d_{\text{tl}}$ to represent distance to an upcoming traffic light and the following constraint when the upcoming traffic light is red at any time during the time interval $[t, t+N_p]$:
\begin{equation} \label{tl}
    0 \leq d_{\text{tl}}(k|t) \,\, \forall\,\,  k=t,...,t+N_p,
\end{equation}
Note that \eqref{tl} enforces the ego vehicle to stop at the red light.

\begin{figure}[!h]
	\centering
 	\includegraphics[width=\linewidth,height=\textheight,keepaspectratio]{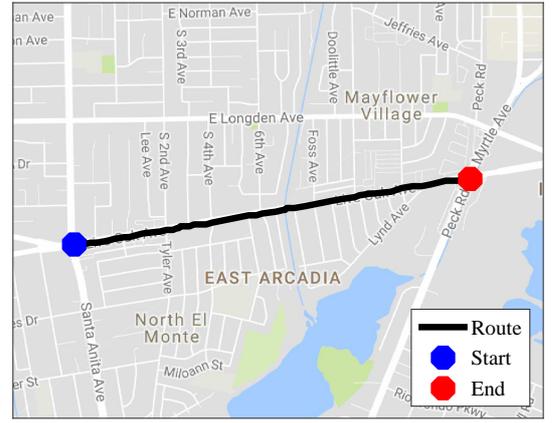}
	\caption{Route in urban road in Arcadia }
	\label{fig:map}
\end{figure}

\begin{figure}[!h]
	\centering
 	\includegraphics[width=\linewidth,height=\textheight,keepaspectratio]{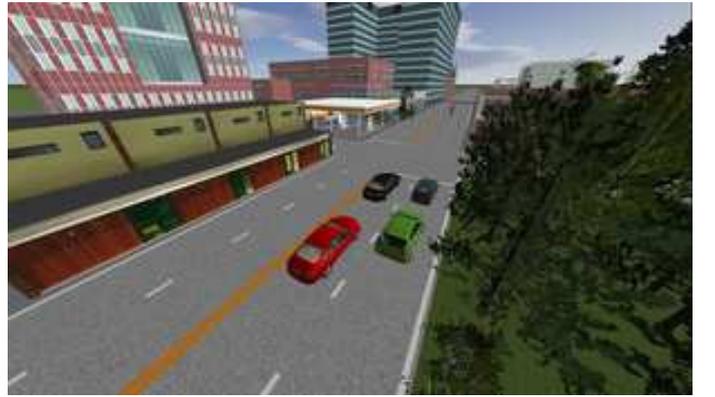}
	\caption{Captured image of PreScan virtual environment of Arcadia route in Figure~\ref{fig:map}}
	\label{fig:urbanmap}
\end{figure}

For this simulation a virtual PreScan and Vissim environment is built based on a real map and actual traffic data collected by Sensys Network \cite{Sensys}.
Specifically, we reproduce an approximately $2.5$km urban road which consists of $8$ traffic signals, located in the city of Arcadia illustrated in Figure~\ref{fig:map}. 
The constructed virtual environment is shown in Figure~\ref{fig:urbanmap}. Our mirrored vehicle in the virtual environment is equipped with a Radar sensor and a lane detection sensor offered by PreScan. We also assume that there is no communication among vehicles; therefore, our vehicle assumes the worst case velocity prediction for the front vehicles which is that the front vehicle can fully decelerate at any moment until it reaches a full stop.
Perfect SPaT information about the upcoming traffic lights is available to the ego vehicle through the Simulink interface. 


For this simulation we use the setting depicted in Figure~\ref{fig:hardware_schematic_simplified} and assume no model mismatch between the longitudinal vehicle dynamics model used in our eco-ACC MPC and that used in Simulink. For lateral dynamics of the vehicle, we implement a simple lane keeping controller using a well-tuned PID \cite{chaib2004h}.

\begin{figure}[!h]
	\centering
 	\includegraphics[width=\linewidth,height=\textheight,keepaspectratio]{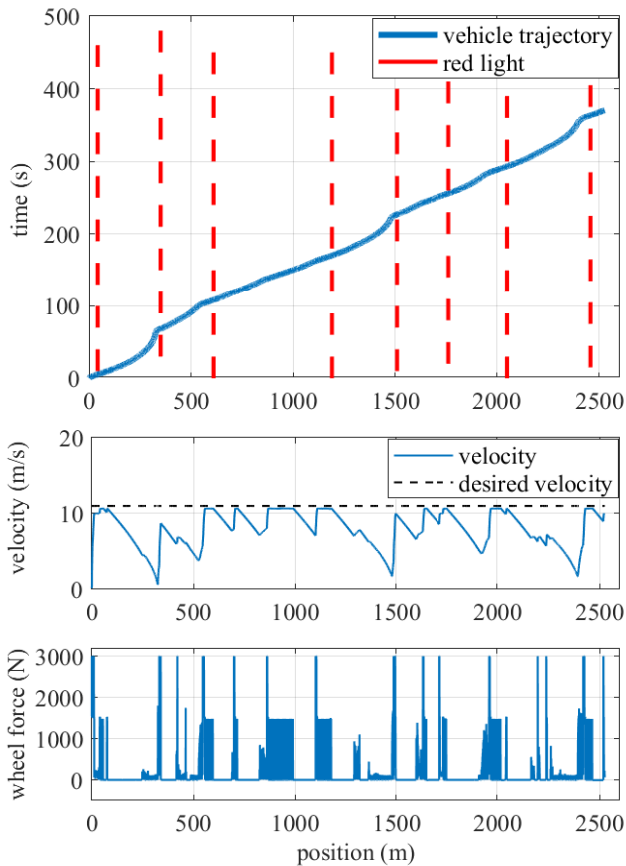}
	\caption{HIL simulation for a vehicle running through a series of simulated traffic lights with other traffic vehicles: first plot shows the vehicle and red light trajectories; second and third plots show the closed loop trajectories of velocity and wheel force applied to the vehicle, respectively}
	\label{fig:urbansim}
\end{figure}

Figure~\ref{fig:urbansim} shows the results of eco-ACC in our HIL urban setting.
The first plot shows the closed loop trajectory, the second plot shows the velocity, and the third plot shows the wheel force with respect to the position.
As seen, our controller continues to switch between applying the accelerating force and coasting without braking. The coasting phase happens early as our vehicle approaches either a red light or the front target vehicles. For instance, immediately after the ego vehicle passes the third traffic light, it starts to coast because of the detected front vehicle.

A few remarks follow:
	\begin{itemize}
	 \item  We successfully tested our controller implemented on an on-board ECU in the virtual urban environment built with real collected traffic data.
	 \item The eco-ACC can maneuver safely in the urban street which involves other traffic vehicles and traffic lights.
	 \item Using the eco-ACC, the ego vehicle only decelerates by coasting, thus minimizing the waste of energy.
	\end{itemize}

\section{Conclusion}\label{sec:conclusion}
In this paper we proposed a hardware-in-the-loop simulation setup appropriate for the development of connected automated vehicles. 
The key aspect of our work is to connect a physics-based simulation platforms such as PreScan and Vissim with the on-board ECUs and the dedicated vehicle dynamics/powertrain simulator. 
We tested an eco-ACC controller with our proposed HIL setting which includes on-board ECUs, a high fidelity vehicle simulator, and the virtual environment which is established based on the real data. 
Future works include developing a more advanced vehicle dynamics and powertrain controller for CAVs by utilizing the connectivity with cloud computing and validating the energy savings in our HIL setting. Furthermore, we can compare our HIL simulation results with those from experiments in real environments.  



\section{Acknowledgement}
The information, data, or work presented herein was funded in part by the Advanced Research Projects Agency-Energy (ARPA-E), U.S. Department of Energy, under Award Number DE-AR0000791. The views and opinions of authors expressed herein do not necessarily state or reflect those of the United States Government or any agency thereof.

The authors would also like to thank to Hyundai Motor Company for providing us with an ETAS simulator and a Dyno facility and Sensys Networks for sharing the traffic data used in this paper.

\printbibliography
\end{document}